\documentclass[conference]{IEEEtran}
\IEEEoverridecommandlockouts
\usepackage[compatibility=false]{caption}
\usepackage{cite}
\usepackage{amsmath,amssymb,amsfonts}
\usepackage{graphicx}
\usepackage{textcomp}
\usepackage{xcolor}
\usepackage{microtype}
\usepackage{inconsolata}
\usepackage{algorithm}
\usepackage{algpseudocode}
\usepackage{float}
\usepackage{multicol}
\usepackage{multirow}
\usepackage{booktabs}
\usepackage{subcaption}
\usepackage{colortbl}
\usepackage{longtable}
\usepackage{enumitem}
\usepackage{etoc}
\usepackage{verbatim}
\usepackage{fancyvrb}
\usepackage{fvextra}
\usepackage{tcolorbox}
\usepackage{hyperref}
\def\BibTeX{{\rm B\kern-.05em{\sc i\kern-.025em b}\kern-.08em
    T\kern-.1667em\lower.7ex\hbox{E}\kern-.125emX}}

\tcbuselibrary{breakable}

\begin{document}

\title{Semantic-Preserving Prompt Hijacking: A Black-Box Adversarial Attack on Auto-Prompt Optimization}
\author{
    \IEEEauthorblockN{Chong Zhang\textsuperscript{\rm *, 1}, 
                      Xiang Li\textsuperscript{\rm *, 2}, 
                      Jia Wang\textsuperscript{\rm 1}, 
                      Shan Liang\textsuperscript{\rm 1}, 
                      Haochen Xue\textsuperscript{\rm 1}, 
                      Xiaobo Jin\textsuperscript{\rm 1$^{\dagger}$} 
                      \thanks{$^{\dagger}$ Corresponding Author. $^{*}$ Equal Contribution.\hfill \break
                      \hspace*{1.2em} This work was partially supported by the “Qing Lan Project” in Jiangsu universities, NSFC (No. 72401232) and RDF-24-01-016.}}
    \IEEEauthorblockA{\textsuperscript{1}Xi'an Jiaotong-Liverpool University, 
                      \textsuperscript{2}The Chinese University of Hong Kong \\
                      \textbf{Email:} Chong.zhang19@student.xjtlu.edu.cn, Xiaobo.jin@xjtlu.edu.cn}
}
\maketitle

\begin{abstract}
LLMs increasingly integrate auto-suggestion optimization modules, enabling them to rewrite and display user input before generating the final response. While this design aims to enhance transparency and trust, its process of autonomously selecting a single ``best" result from multiple candidate solutions allows attackers to hijack this optimization process by inducing subtle, imperceptible semantic shifts. To address this, we propose a semantic preservation hijacking attack method based on black-box conditions—Adaptive Greedy Local Search. This method hierarchically decomposes the input text, masks key language units, and dynamically adjusts candidate replacement words at predefined semantic checkpoints. This maximizes the deviation between the model output and the original intent while strictly maintaining semantic similarity to the original text. Experimental results on commercial and open-source LLMs demonstrate that, under the same semantic similarity constraints, this method achieves a higher attack success rate than existing attack methods in over 2400 test cases. Code is available at: \url{https://github.com/franz-chang/Adaptive_Greedy_Local_Search}.
\end{abstract}


\section{Introduction}
\label{sec:intro}

Large language models (LLMs) are increasingly being used in conjunction with automatic suggestion optimization (APO) modules, which explicitly rewrite user input before generating a response. Some well-known commercial systems, such as Microsoft Bing Copilot~\cite{bingcopilot2023}, Claude.ai's automatic expansion feature~\cite{claude2024}, and Perplexity Pro's real-time restatement feature~\cite{perplexity2024}, display "improved" versions of user suggestions to enhance transparency and usability. While these optimizers aim to build user trust by revealing their modifications, their underlying selection mechanism—autonomously choosing a "best" candidate from multiple alternatives—introduces a new avenue for attack. Attackers can cleverly scramble the original input to induce semantic drift, thereby hijacking the optimizer's selection process without the user's notice.

Previous research on suggestion injection and adversarial attacks~\cite{zou2023universal} has primarily focused on scenarios where the optimization process is opaque or entirely internal. However, the visibility of optimization suggestions presents a unique challenge: attacks must maintain semantic consistency to appear legitimate while also making meaningful changes to the model's output. Existing black-box adversarial methods, such as beam search-based attacks ~\cite{choi2022tabs,zhu2023beamattack}, often lack the dynamic adaptability needed to simulate real-world optimizer behavior under strict similarity constraints, leading to low attack success rates or detectable perturbations.
\begin{figure}
\centering
\includegraphics[width=1.0\linewidth]{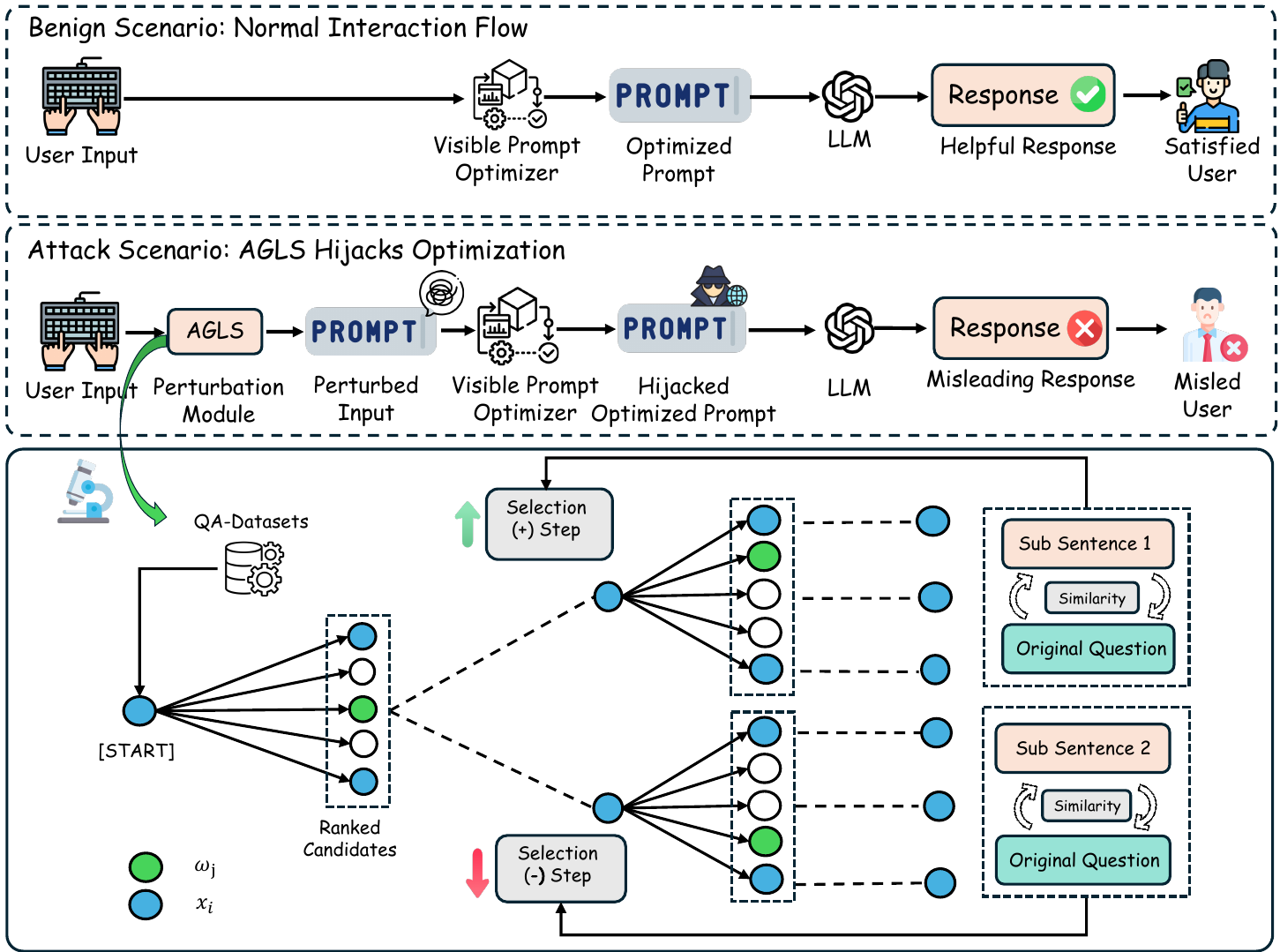}
\caption{Motivation: Semantic Hijacking Attack on Visible Prompt Optimizers in LLMs}
\label{fig:intro}
\end{figure}

To address this deficiency, we propose Adaptive Greedy Local Search (AGLS), a black-box attack framework that leverages visible-cue optimizers. AGLS operates by decomposing the input text into hierarchical clauses, masking key language units, and dynamically adjusting candidate replacements at semantic checkpoints. This allows the attack to maximize the behavioral differences of the target language logic model (LLM) while maintaining high semantic similarity to the original input. We evaluate AGLS on multiple commercial and open-source language learning models (GPT-4, Llama-3, Qwen-2.5, Gemma-2) for different question-answering and inference datasets. Experimental results show that, under similar similarity budgets, AGLS consistently outperforms state-of-the-art adversarial methods, achieving a higher attack success rate across over 2400 test cases. Our contributions are primarily in the following three aspects:
\begin{itemize}
    \item We identify and formalize the novel attack surface caused by visible cue optimization modules in production-level language learning models.
    \item We propose AGLS, a novel black-box attack method that dynamically balances semantic preservation and attack effectiveness through adaptive checkpoint-based search.
    \item We provide extensive empirical validation and ablation experiments, along with practical guidelines for designing more robust cue optimization systems.
\end{itemize}

\section{Related Work}
\subsection{Adversarial Attacks for Large Language Models}

Adversarial attacks on LLMs have been extensively studied in both white-box and black-box scenarios. White-box methods, such as GBDA \cite{guo2021gradientbased}, HotFlip \cite{ebrahimi2018hotflip}, AutoPrompt \cite{shin2020autoprompt}, and universal adversarial triggers \cite{wallace2021universal}, leverage gradient information to craft perturbations but are inherently limited to accessible model architectures, rendering them ineffective against closed-source commercial systems. In contrast, black-box attacks operate without internal knowledge and primarily employ strategies such as token-level substitution (e.g., BERT-Attack \cite{li2020bert}), semantically equivalent adversaries (SEAs \cite{ribeiro2018semantically}), prompt injection \cite{clusmann2024prompt,greshake2023not}, query-free or clean-label backdoor attacks \cite{salem2022dynamic,yan2024backdooring}, model stealing \cite{papernot2017practical}, and goal-driven optimization methods \cite{zhang2025customized} that maximize statistical divergence measures such as KL divergence or Mahalanobis distance \cite{zhang2024goal,zhang2024target}. Collectively, these approaches expose a broad spectrum of security vulnerabilities in modern LLM systems \cite{wang2024large}. However, most existing attacks assume direct model access and do not account for the increasingly prevalent intermediate processing layers—such as visible prompt optimizers—that reshape user inputs before they reach the core model.

\subsection{Beam Search in Adversarial Text Generation}

Beam search has been widely adopted to improve the coherence and semantic consistency of adversarial examples. TABS \cite{choi2022tabs} introduces a tree-based adversarial beam search mechanism that maintains a set of Top-$k$ candidates while incorporating contextual semantic constraints to narrow the search space. Subsequent improvements include backtracking strategies to expand candidate diversity \cite{zhao2021generating}, as well as methods that leverage transferability and random word replacement prior to beam selection \cite{zhu2022leveraging}. BeamAttack \cite{zhu2023beamattack} further integrates semantic filtering to retain the most similar replacements, thereby enhancing the quality of adversarial samples \cite{wang2024large}. A common limitation across these methods is their reliance on static or precomputed semantic filtering, which often leads to \textit{early-commitment bias}—wherein unpromising candidates are pruned prematurely, necessitating costly resampling if later stages require alternative options. This rigidity becomes particularly problematic in dynamic, multi-stage settings such as prompt optimization pipelines, where semantic preservation must be continuously monitored.

\section{Methodology}

\subsection{Problem Formulation}

Given an input text $t$ consisting of multiple sentences, our goal is to generate an adversarial text $t'$ that can mislead the target Large Language Model (LLM) while preserving the semantic meaning of the original input. We denote the target LLM as $O$, and its outputs for $t$ and $t'$ as $r$ and $r'$, respectively. The semantic similarity between $t$ and $t'$ is measured by the function $S(t, t')$. A successful attack must satisfy the following:
\begin{equation}
O(t) = r, \quad O(t') \neq r, \quad S(t, t') \geq \sigma,
\end{equation}
where $\sigma$ is a predefined similarity threshold. This form ensures that the adversarial sample $t'$ is both effective (causing misclassification) and difficult to detect (semantically similar to $t$.

\subsection{Adaptive Greedy Local Search (AGLS)}

\begin{figure*}
    \centering
    \includegraphics[width=1.0\linewidth]{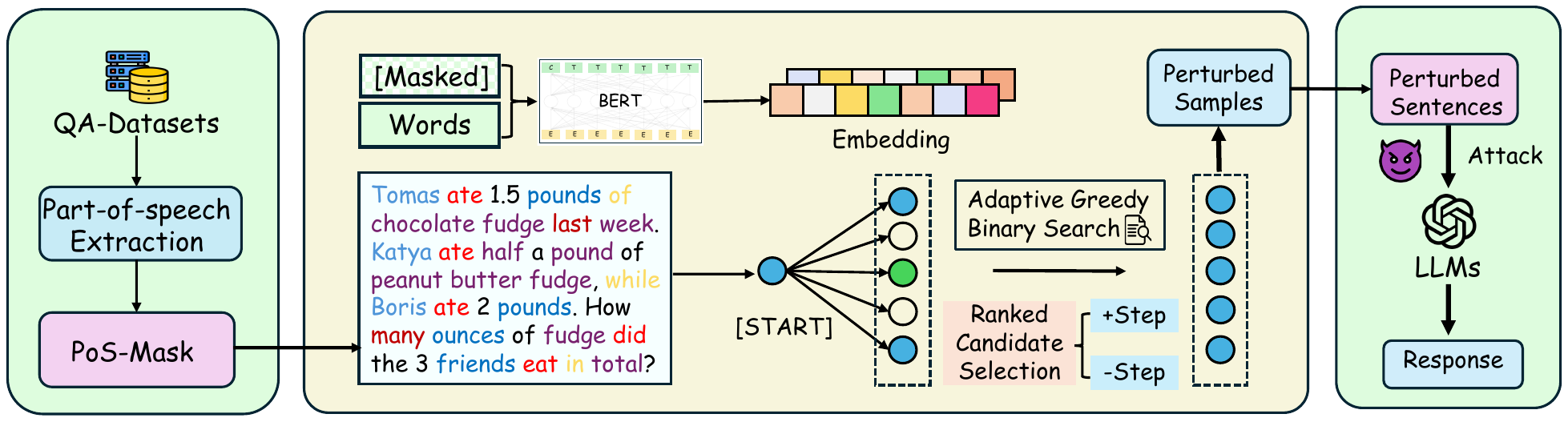}
    \caption{Overall framework of the proposed Adaptive Greedy Local Search (AGLS) attack: \textbf{(a).} PoS (Part-of-Speech) extraction and masking. \textbf{(b).} Generation of AGLS perturbed samples. \textbf{(c).} Perturbed Samples Attack.}
    \label{fig:framework}
\end{figure*}

To construct adversarial examples under the above constraints, we propose \textbf{Adaptive Greedy Local Search (AGLS)}, a black-box attack method that dynamically adjusts lexical substitutions based on real-time semantic feedback. The overall framework is shown in Figure~\ref{fig:framework}, comprising three main stages: (a) part-of-speech tagging and keyword masking, (b) iterative candidate word generation with adaptive control, and (c) adversarial example verification.

\subsubsection{Input Decomposition and Checkpoint Setting}

Given an input sentence $X = \{x_1, x_2, \dots, x_T\}$, we first segment it into $n$ clauses based on punctuation and syntactic boundaries. The end of each clause is designated as a \emph{checkpoint} $t_i$, where $t_1 < t_2 < \cdots < t_n$. At each checkpoint $t_i$, we compute the semantic similarity between the original prefix $X_{1:t_i}$ and the partially generated adversarial prefix $\hat{X}_{1:t_i}$, where \begin{equation}
X_{1:t_i} = \{x_1, x_2, \dots, x_{t_i}\},
\end{equation}
and 
\begin{equation}
\hat{X}_{1:t_i} = \{\hat{x}_1, \hat{x}_2, \dots, \hat{x}_{t_i}\}. 
\end{equation}

\subsubsection{Dynamic Candidate Word Adjustment}

At checkpoint $t_i$, we maintain a candidate word set $\mathcal{C}(t_i) = \{c_1, c_2, \dots, c_k\}$, which is sorted by the masked language model (BERT). The similarity between the original text and the generated text is calculated as follows:
\begin{equation}
\sigma_{\text{sim}}^{(i)} = \text{SIM}\bigl(X_{1:t_i}, \hat{X}_{1:t_i}\bigr),
\end{equation}
where, \textbf{SIM} represents the cosine similarity between BERT word embeddings. A dynamic threshold $\sigma_{\text{th}}$ is used to guide the selection of candidate words $c_p$. The adjustment rules are as follows: 

\begin{align}
\text{If } \sigma_{\text{sim}}^{(i)} < \sigma_{\text{th}}, \quad &c_p \leftarrow c_{j - s} \quad \text{(Increase similarity)}, \\
\text{If } \sigma_{\text{sim}}^{(i)} > \sigma_{\text{th}}, \quad &c_p \leftarrow c_{j + s} \quad \text{(Increase deviation)},
\end{align}
where $s$ is a fixed step size. This feedback mechanism ensures that the generated text, while gradually deviating from the attack target, remains within the desired semantic neighborhood.

\subsubsection{Masked Token Generation}

For each masked position $m_i$ corresponding to a keyword (verbs or plural nouns), we use BERT to predict a set of candidate tokens. The probability distribution over the vocabulary is:
\begin{equation}
    P(c \mid X_{\text{M}}) = \frac{\exp(\operatorname{logits}(c))}{\sum_{c' \in \mathcal{C}} \exp(\operatorname{logits}(c'))},
\end{equation}
where $X_{\text{M}}$ is the masked sentence. The top-k candidates are retained, and the final token $\hat{x}_{m_i}$ is selected according to the adaptive rule described above.

\subsubsection{Iterative Sentence Reconstruction}

The adversarial sentence is constructed incrementally. After each substitution, the context is updated as:
\begin{equation}
    X_{\text{M}} \leftarrow \{x_1, \dots, x_{m_i-1}, \hat{x}_{m_i}, x_{m_i+1}, \dots, x_T\}.
\end{equation}
The process repeats until all masked positions are filled. The final adversarial example is denoted as $\hat{X}_{\text{F}}$, and its generation can be summarized as:
\begin{equation}
    \hat{X}_{\text{F}} = \prod_{i=1}^{n} P\Bigl(\hat{X}_{t_i} \mid X_{\text{M}}, \hat{X}_{1:t_{i-1}}, \sigma_{\text{sim}}^{(i)}\Bigr).
\end{equation}

\subsection{Attack Constraints and Objectives}
The proposed Adaptive Greedy Local Search (AGLS) method is designed to satisfy two key requirements:
\begin{itemize}
    \item \textbf{Effectiveness} The adversarial text $t'$ must induce a different model output than the original text $t$, i.e., $O(t') \neq O(t)$.
    \item \textbf{Stealthiness} The semantic similarity $S(t, t')$ must remain above $ \sigma$, ensuring that perturbations are minimal and contextually natural. This is achieved through the adaptive control mechanism that continuously monitors and adjusts the semantic drift.
\end{itemize}

\subsection{Discussions}

\noindent \textbf{Theoretical Basis}  Adversarial attacks that typically preserve semantics solve the following optimization problem
\begin{equation}
    \max_{\delta} \mathcal{D}(O(t + \delta),O(t)),\quad s.t. \quad S(t,t + \delta) \ge \sigma.
\end{equation}

However, for a sentence of length L, if there are $V$ candidate replacement words at each position, the search space size is $O(V^L)$. Furthermore, under the black-box setting, the objective function $\mathcal{D}$ is highly non-convex and non-differentiable, while the constraints are also discrete and non-convex.

Therefore, the core innovation of AGLS lies in decomposing the global constraint optimization into a sequential local decision problem.
\begin{equation}
    \max_{\delta_i} \mathcal{D}(O(t_{i} + \delta_i),O(t_i)),\quad s.t. \quad S(t_i,t_i + \delta_i) \in [\sigma_{\text{low}}^i,\sigma_{\text{high}}^i],
\end{equation}
where $i = 1, 2, \cdots, n$ represents the checkpoint index.

\noindent \textbf{Algorithm Rationality}  This  ``divide and conquer" strategy is theoretically reasonable: it draws on the ideas of dynamic programming and model predictive control, approximating the global optimal solution through local optimal decisions, while avoiding the accumulation of semantic deviations through real-time feedback.

\noindent \textbf{Realism of the Black-Box Assumption}  AGLS does not rely on the gradient or architectural information of the target LLM, requiring only input/output query interfaces (provided by all commercial APIs) and a public, powerful semantic model (BERT). This allows the method to be directly applied to closed-source systems such as GPT-4, Claude, and Gemini, aligning with real-world attack scenarios.

\noindent \textbf{Controllable Computational Cost} Compared to attack methods based on reinforcement learning or large-scale sampling, AGLS's greedy search strategy keeps the time complexity to $O(n\times k)$, where $n$ is the number of checkpoints and $k$ is the candidate pool size. The average generation time ($\sim$ 2 seconds, see Table \ref{tab:main-test}) in experiments is feasible in interactive attack scenarios.

\section{Experiments}

\subsection{Experimental Setup}
\label{sec:setup}

\noindent \textbf{Datasets} We evaluate AGLS on four benchmark datasets spanning diverse reasoning and comprehension tasks:
\begin{itemize}
    \item \textbf{GSM8K}~\cite{cobbe2021training}: A dataset of grade-school math word problems requiring multi-step reasoning.
    \item \textbf{Math QA}~\cite{amini2019mathqa}: A dataset of math problems with operation-based formalisms, testing numerical reasoning.
    \item \textbf{SQuAD}~\cite{rajpurkar2016squad}: A reading comprehension dataset where answers are extracted from Wikipedia paragraphs.
    \item \textbf{Strategy QA}~\cite{geva2021didaristotleuselaptop}: A question-answering benchmark requiring implicit multi-hop reasoning.
\end{itemize}
These datasets collectively assess an attack's effectiveness across both \textit{numeric} and \textit{textual} response scenarios.
\\
\noindent \textbf{Victim Models} We test commercial and open-source LLMs
\begin{itemize}
    \item \textbf{Commercial:} GPT-4~\cite{achiam2023gpt} and GPT-4o (latest API versions).
    \item \textbf{Open-source:} Llama-3.1 series (8B, 70B)~\cite{dubey2024llama}, Llama-3.2 series (3B)~\cite{dubey2024llama}, Qwen2.5 series (7B, 14B)~\cite{yang2024qwen2}, and Gemma-2 series (9B, 27B)~\cite{team2024gemma}.
\end{itemize}
This selection ensures coverage of models with varying scales, architectures, and alignment strategies.

\noindent \textbf{Evaluation Metrics} Let $D$ be the test set and $T \subseteq D$ the subset where the victim model correctly answers the clean input $x$ with ground truth $y$. For an adversarial example $a(x)$ generated from $x$, we report:
\begin{itemize}
    \item \textbf{Clean Accuracy ($\mathcal{A}_{\text{clean}}$):} Model accuracy on original inputs: $\frac{|T|}{|D|}$.
    \item \textbf{Attack Accuracy ($\mathcal{A}_{\text{attack}}$):} Model accuracy on adversarial inputs: $\frac{|\{(x,y)\in T \mid f(a(x)) = y\}|}{|D|}$.
    \item \textbf{Attack Success Rate (ASR):} Proportion of successfully attacked samples: $\text{ASR} = 1 - \frac{\mathcal{A}_{\text{attack}}}{\mathcal{A}_{\text{clean}}}$.
    \item \textbf{Average Inference Time ($\text{T}_{\text{avg}}$):} Average time to generate an adversarial sample.
\end{itemize}
Notably, ASR isolates the attack's effectiveness from the model's inherent accuracy, providing a cleaner measure of adversarial vulnerability.

\begin{table*}[!ht]
\centering
\caption{Attack performance of AGLS across different LLMs and datasets, including numerical and text response tasks}
\label{tab:main-test}
\scriptsize
\begin{tabular}{lcccccccccccccccc}
\toprule
\multirow{2}{*}{\textbf{Models}}
& \multicolumn{4}{c}{\textbf{GSM8K}}
& \multicolumn{4}{c}{\textbf{Math QA}}
& \multicolumn{4}{c}{\textbf{SQuAD}}
& \multicolumn{4}{c}{\textbf{Strategy QA}} \\
\cmidrule(lr){2-5} \cmidrule(lr){6-9} \cmidrule(lr){10-13} \cmidrule(lr){14-17}
& $A_{\textrm{clean}}$ & $A_{\textrm{attack}}$ & \textbf{ASR}↑ & $\text{T}_{\text{avg}}$(s)
& $A_{\textrm{clean}}$ & $A_{\textrm{attack}}$ & \textbf{ASR}↑ & $\text{T}_{\text{avg}}$(s)
& $A_{\textrm{clean}}$ & $A_{\textrm{attack}}$ & \textbf{ASR}↑ & $\text{T}_{\text{avg}}$(s)
& $A_{\textrm{clean}}$ & $A_{\textrm{attack}}$ & \textbf{ASR}↑ & $\text{T}_{\text{avg}}$(s) \\
\midrule
gpt-4o-latest & 47.50 & 15.00 & \cellcolor[gray]{0.9}68.42   & 2.21 & 32.40 & 22.33 & \cellcolor[gray]{0.9}\textbf{31.08} & 0.82 & 54.52 & 43.49 & \cellcolor[gray]{0.9}20.23         & 0.63 & 55.33 & 43.45 & \cellcolor[gray]{0.9}21.47         & 0.70 \\
gpt-4   & 27.50 & 2.50  & \cellcolor[gray]{0.9}\textbf{90.91} & 2.31 & 48.50 & 34.50 & \cellcolor[gray]{0.9}28.87          & 1.12 & 51.84 & 32.43 & \cellcolor[gray]{0.9}\textbf{37.44} & 0.78 & 57.33 & 43.70 & \cellcolor[gray]{0.9}\textbf{23.77} & 0.56 \\
\hline
llama3.1-8B   & 17.50 & 7.50  & \cellcolor[gray]{0.9}57.14         & 1.49 & 8.67  & 4.67  & \cellcolor[gray]{0.9}46.12         & 0.23 & 33.78 & 30.43 & \cellcolor[gray]{0.9}9.92          & 0.56 & 43.33 & 42.33 & \cellcolor[gray]{0.9}2.31          & 1.07 \\
llama3.1-70B  & 47.50 & 5.00  & \cellcolor[gray]{0.9}\textbf{89.47} & 1.72 & 13.67 & 11.67 & \cellcolor[gray]{0.9}17.12         & 1.27 & 47.83 & 27.33 & \cellcolor[gray]{0.9}\textbf{42.86}         & 1.13 & 47.33 & 35.68 & \cellcolor[gray]{0.9}\textbf{24.61} & 1.42 \\
llama3.2-3B   & 47.50 & 5.00  & \cellcolor[gray]{0.9}89.47         & 1.38 & 3.00  & 1.00  & \cellcolor[gray]{0.9}\textbf{66.67} & 0.15 & 28.09 & 22.41 & \cellcolor[gray]{0.9}20.22  & 0.36 & 34.33 & 29.33 & \cellcolor[gray]{0.9}14.56  & 0.23 \\
\hline
qwen2.5-7B    & 15.00 & 7.50  & \cellcolor[gray]{0.9}50.00         & 1.91 & 10.33 & 2.00  & \cellcolor[gray]{0.9}\textbf{80.64} & 0.69 & 20.74 & 19.40 & \cellcolor[gray]{0.9}\textbf{6.46}          & 0.72 & 55.00 & 36.57 & \cellcolor[gray]{0.9}\textbf{33.51} & 0.46 \\
qwen2.5-14B   & 22.50 & 5.00  & \cellcolor[gray]{0.9}\textbf{77.78} & 2.49 & 64.86 & 39.18 & \cellcolor[gray]{0.9}39.59  & 1.16 & 30.77 & 30.43 & \cellcolor[gray]{0.9}1.10  & 1.12 & 55.67 & 44.43 & \cellcolor[gray]{0.9}20.19 & 0.68 \\
\hline
gemma2-9B   & 12.50 & 7.50  & \cellcolor[gray]{0.9}40.00  & 1.24 & 7.67  & 3.67  & \cellcolor[gray]{0.9}52.15         & 0.44 & 34.11 & 27.42 & \cellcolor[gray]{0.9}\textbf{19.61}         & 0.45 & 54.00 & 48.36 & \cellcolor[gray]{0.9}10.44         & 0.17 \\
gemma2-27B    & 70.00 & 10.00 & \cellcolor[gray]{0.9}\textbf{85.71} & 0.85 & 10.33 & 2.00  & \cellcolor[gray]{0.9}\textbf{80.64} & 1.28 & 44.15 & 37.79 & \cellcolor[gray]{0.9}14.41         & 0.85 & 64.00 & 59.67 & \cellcolor[gray]{0.9}6.77          & 0.21 \\
\bottomrule
\end{tabular}
\end{table*}

\subsection{Implementation Details}
All experiments are conducted in a \textit{strictly black-box} setting. We use \texttt{bert-base-uncased} as both the masked language model for candidate generation and the sentence encoder for similarity computation (768-dimensional embeddings). Key hyperparameters are fixed: semantic similarity threshold $\sigma = \sigma_{\text{th}} = 0.80$ (BERTScore F1), no length normalization ($\alpha = 1.0$), initial candidate pool drawn from the top-13,000 BERT predictions (covering $>99.9\%$ of probable tokens), final candidate set size $k = 100$, and beam width of 4. The rank-adjustment step size $s = 50$ is triggered only when the BERTScore deviation exceeds 0.15 for three consecutive checkpoints. Each model is evaluated on 1,000 randomly sampled question-answer pairs per dataset.

\subsection{Main Results}

\noindent \textbf{Overall Attack Performance} The main results are summarized in Table~\ref{tab:main-test}. AGLS demonstrates strong and consistent attack performance across both numeric and textual QA tasks. Several key observations emerge:
\begin{itemize}
    \item \textbf{High ASR on Numerical Reasoning:} On GSM8K and Math QA, AGLS achieves high ASR (often $>60\%$) against most models, including GPT-4o and Llama-70B. This indicates that numerical reasoning, while structured, is highly sensitive to subtle semantic perturbations in the problem formulation.
    \item \textbf{Model-Scale Sensitivity:} Larger models (e.g., Llama-3.1-70B) generally show higher clean accuracy but remain vulnerable, with ASR up to 89.47\% on GSM8K. This suggests that robustness does not scale linearly with parameters, and advanced optimization modules can become attack vectors regardless of model capacity.
    \item \textbf{Lower ASR on Textual QA:} For SQuAD and StrategyQA, ASR is generally lower than on math datasets. This aligns with the intuition that factual and commonsense reasoning may rely on more distributed semantic representations, making them slightly more resilient to localized token substitutions.
    \item \textbf{Efficiency:} The average generation time ($\text{T}_{\text{avg}}$) ranges from 0.17s to 2.49s, confirming the practical feasibility of AGLS for near-real-time attacks.
\end{itemize}
A notable exception is the Gemma-2-27B model, which shows relatively low ASR on textual QA, potentially due to its strong instruction-tuning or inherent robustness properties—a direction for future analysis.

\begin{table*}[!ht]
\centering
\caption{Comparison of AGLS with state-of-the-art adversarial attack methods on GPT-3.5-turbo. Note that a higher ASR (Attack Success Rate \%) is better. }
\footnotesize
\label{tab:main-com}
\begin{tabular}{lccccccccc}
\toprule
\multirow{2}{*}{\textbf{Models}} & \multicolumn{3}{c}{\textbf{SQuAD2.0}} & \multicolumn{3}{c}{\textbf{Math}} & \multicolumn{3}{c}{\textbf{SVAMP}} \\
\cmidrule(lr){2-4} \cmidrule(lr){5-7} \cmidrule(lr){8-10}
& $\mathcal{A}_{\textrm{clean}}$ & $\mathcal{A}_{\textrm{attack}}$ & \textbf{ASR} & $\mathcal{A}_{\textrm{clean}}$ & $\mathcal{A}_{\textrm{attack}}$ & \textbf{ASR}  & $\mathcal{A}_{\textrm{clean}}$ & $\mathcal{A}_{\textrm{attack}}$ & \textbf{ASR} \\
\midrule
BertAttack \cite{li2020bert} & 71.16 & 24.67 & 65.33 & 72.30 & 44.82 & 38.01 & 88.00 & 77.41 & 12.03  \\ 
DeepWordBug \cite{gao2018black} & 71.16 & 65.68 & 6.72 & 72.30 & 48.36 & 33.11 &  88.00 & 64.83 & 26.33   \\
TextFooler \cite{jin2020bert} & 71.16 & 15.60 & 78.59 & 72.30 & 46.80 & 35.27 &  88.00 & 43.62 & 50.43  \\
TextBugger \cite{li2018textbugger} & 71.16 & 60.14 & 16.08 & 72.30 & 47.75 & 33.96 &  88.00 & 60.72 & 20.77  \\
Stress Test \cite{ribeiro2020beyond} & 71.16 & 70.66 & 1.78 & 72.30 & 39.59 & 45.24 &  88.00 & - & -  \\
CheckList \cite{ribeiro2020beyond} & 71.16 & 68.81 & 3.64 & 72.30 & 36.90 & 48.96 &  88.00 & - & -  \\
G2PIA \cite{zhang2024goal} & 71.16 & 14.00 & 79.50 & 72.30 & 52.37 & 27.57 & 88.00 & 69.42 & 21.11 \\
Target-driven Attack \cite{zhang2024target}  & 71.16 & 14.91 & 83.02  & 72.30 & 33.39 & 53.82 &  88.00 & 64.87 & 20.28 \\
BeamAttack \cite{zhu2023beamattack} & 71.16 & 48.33 & 32.08 & 72.30 & 52.25 & 27.73 & 88.00 & 47.50 & 46.02\\
\midrule
\rowcolor{gray!20} 
\textbf{Our Method} & 71.16 & 12.48 & \textbf{83.09} & 72.30 & 28.50 & \textbf{60.61} &  88.00 & 33.33 & \textbf{62.13} \\
\bottomrule 
\end{tabular}
\end{table*}

\subsection{Comparison with State-of-the-Art Attacks}
\label{sec:comparison}
We compare AGLS against seven prominent adversarial attack methods on GPT-3.5-turbo, following the evaluation protocol of PromptBench~\cite{zhu2024promptbench}. Table~\ref{tab:main-com} presents the results on SQuAD 2.0, Math, and SVAMP datasets. Key Findings are as follows:
\begin{itemize}
    \item AGLS achieves the \textbf{highest ASR} on all three datasets, outperforming strong baselines like TextFooler~\cite{jin2020bert}, BERT-Attack~\cite{li2020bert}, and the recent G2PIA~\cite{zhang2024goal} and TDA~\cite{zhang2024target}. For instance, on the Math dataset, AGLS attains an ASR of 60.61\%, a 6.79-point improvement over the second-best method (TDA at 53.82\%).
    \item Unlike methods that rely on extensive querying (e.g., beam search variants), AGLS maintains a balanced query count while achieving superior effectiveness. This highlights the efficiency of its adaptive feedback mechanism.
    \item The significantly lower \textit{attack accuracy} (\(\mathcal{A}_{\text{attack}}\)) of AGLS—e.g., 12.48\% vs. 48.33\% for BeamAttack on SQuAD 2.0—directly translates to its higher ASR, confirming its ability to more reliably flip correct predictions.
\end{itemize}
These results validate AGLS as a new state-of-the-art method for semantic-preserving adversarial attacks, particularly in scenarios mimicking real-world prompt optimization pipelines.

\subsection{Ablation Studies}
\label{sec:ablation}

\noindent \textbf{Hyperparameter Sensitivity} We analyze the sensitivity of AGLS to two key hyperparameters: the global similarity threshold $\sigma$ and the beam-width adjustment ratio $\omega$. Table~\ref{tab:hyper_sen} shows results on the GSM8K dataset across four representative models. The combination $\sigma = 0.8, \omega = 0.7$ consistently yields the highest ASR. Lowering $\sigma$ to 0.3 increases attack accuracy ($\mathcal{A}_{\text{attack}}$) but at the cost of reduced ASR, as perturbations become more detectable (semantic similarity drops). Conversely, setting $\omega = 0.3$ (narrower beam) often leads to ASR degradation, confirming that a broader candidate pool is beneficial for finding effective perturbations.

\begin{table}[!ht]
    \centering
    \caption{Hyperparameter sensitivity analysis for $\sigma$ and $\omega$ on GSM8K.}
    \label{tab:hyper_sen}
    \footnotesize
    \begin{tabular}{l|ccccc}
        \toprule
        \textbf{Target Models} & $\sigma$ & $\omega$ & $\mathcal{A}_{\textrm{clean}}$ & $\mathcal{A}_{\textrm{attack}}$ & \textbf{ASR} \\
        \midrule
                                    & 0.3 & 0.7 & 17.50 & 8.50 & \cellcolor[gray]{0.9}51.43 \\
        llama3.1-8B                 & \underline{0.8} & \underline{0.7} & 17.50 & 7.50 & \cellcolor[gray]{0.9}\textbf{57.14} \\
                                    & 0.8 & 0.3 & 17.50 & 15.00 & \cellcolor[gray]{0.9}14.29 \\
        \midrule
                                    & 0.3 & 0.7 & 15.00 & 9.50 & \cellcolor[gray]{0.9}36.67 \\
        Qwen2.5-7B                  & \underline{0.8} & \underline{0.7} & 15.00 & 7.50 & \cellcolor[gray]{0.9}\textbf{50.00} \\
                                    & 0.8 & 0.3 & 15.00 & 15.00 & \cellcolor[gray]{0.9}0.00 \\
        \midrule
                                    & 0.3 & 0.7 & 50.00 & 15.00 & \cellcolor[gray]{0.9}70.00 \\
        Gemma2-9B                   & \underline{0.8} & \underline{0.7} & 50.00 & 5.00 & \cellcolor[gray]{0.9}\textbf{90.00} \\
                                    & 0.8 & 0.3 & 50.00 & 17.50 & \cellcolor[gray]{0.9}65.00 \\
        \midrule
                                    & 0.3 & 0.7 & 27.50 & 12.50 & \cellcolor[gray]{0.9}54.55 \\
        gpt-4-0125-preview                      & \underline{0.8} & \underline{0.7} & 27.50 & 2.50 & \cellcolor[gray]{0.9}\textbf{90.91} \\
                                    & 0.8 & 0.3 & 27.50 & 9.75 & \cellcolor[gray]{0.9}64.55 \\
        \bottomrule
    \end{tabular}
\end{table}

\noindent \textbf{Dynamic vs. Static Search Strategy} We ablate the core adaptive mechanism by comparing our dynamic candidate adjustment against a static strategy that always selects the median-ranked candidate ($c_{\lfloor k/2 \rfloor}$). Results in Table~\ref{tab:able_dyn} show that the dynamic strategy achieves significantly higher ASR across all models—e.g., improving from 36.36\% to 90.91\% for GPT-4. This validates the necessity of real-time semantic feedback for balancing attack potency and stealth.

\begin{table}[!ht]
    \centering
    \footnotesize
    \caption{Effect of dynamic candidate adjustment vs. static median selection on GSM8K.}
    \label{tab:able_dyn}
    \begin{tabular}{l|l|ccc}
        \toprule
        \textbf{Type} & \textbf{Target Models} & $\mathcal{A}_{\textrm{clean}}$ & $\mathcal{A}_{\textrm{attack}}$ & \textbf{ASR} \\
        \midrule
        \textbf{Dynamic}   & llama3.1-8B            & 17.50 & 7.50 & \cellcolor[gray]{0.9}\textbf{57.14} \\
        Static             & llama3.1-8B            & 17.50 & 15.00 & \cellcolor[gray]{0.9}14.23 \\
        \midrule
        \textbf{Dynamic}   & llama3.2-3B            & 47.50 & 5.00 & \cellcolor[gray]{0.9}\textbf{89.47} \\
        Static             & llama3.2-3B            & 47.50 & 27.75 & \cellcolor[gray]{0.9}41.58 \\
        \midrule
        \textbf{Dynamic}   & qwen2.5-7B             & 15.00 & 7.50 & \cellcolor[gray]{0.9}\textbf{50.00} \\
        Static             & qwen2.5-7B             & 15.00 & 12.50 & \cellcolor[gray]{0.9}16.67 \\
        \midrule
        \textbf{Dynamic}   & gpt-4-0125-preview           & 27.50 & 2.50 & \cellcolor[gray]{0.9}\textbf{90.91} \\
        Static             & gpt-4-0125-preview            & 27.50 & 17.50 & \cellcolor[gray]{0.9}36.36 \\
        \bottomrule
    \end{tabular}
\end{table}

\begin{table*}[ht]
\centering
\footnotesize
\caption{Attack success rate with varying AGLS search scope (candidate pool size).}
\label{tab:beam_width_trend_2}
\begin{tabular}{l|c|ccc|ccc|ccc}
\toprule
\multirow{2}{*}{\textbf{Models}} & \multirow{2}{*}{\textbf{Search Scope}} 
 & \multicolumn{3}{c}{\textbf{GSM8K}} & \multicolumn{3}{c}{\textbf{SQUAD}} & \multicolumn{3}{c}{\textbf{SVAMP}} \\
\cmidrule(lr){3-5} \cmidrule(lr){6-8} \cmidrule(lr){9-11}
 & & $\mathcal{A}_{\textrm{clean}}$ & $\mathcal{A}_{\textrm{attack}}$ & $\textbf{ASR}$ $\uparrow$ 
 & $\mathcal{A}_{\textrm{clean}}$ & $\mathcal{A}_{\textrm{attack}}$ & $\textbf{ASR}$ $\uparrow$ 
 & $\mathcal{A}_{\textrm{clean}}$ & $\mathcal{A}_{\textrm{attack}}$ & $\textbf{ASR}$ $\uparrow$ \\
\midrule
\multirow{5}{*}{\textbf{Llama3.2-3B}} 
 & 2000  & 55.00 & 10.00 & 81.82 & 26.76 & 22.07 & 17.53 & 43.00 &  8.67 & 79.84 \\
 & 6000  & 47.50 &  2.50 & 94.74 & 26.09 & 23.08 & 11.54 & 40.33 &  7.00 & 82.64 \\
 & 10000 & 25.00 &  2.50 & 90.00 & 25.08 & 21.74 & 13.32 & 38.00 &  7.33 & 80.71 \\
 & 13000 & \textbf{42.50} & \textbf{2.50} & \textbf{94.12} & \textbf{25.42} & \textbf{17.48} & \textbf{31.24} & \textbf{39.00} & \textbf{6.00} & \textbf{84.62} \\
 & 16000 & 40.00 &  2.50 & 93.75 & 25.08 & 21.40 & 14.67 & 36.00 &  7.33 & 79.64 \\
\hline
\multirow{5}{*}{\textbf{Qwen2.5-14B}} 
 & 2000  & 10.00 &  7.50 &  2.50 & 32.78 & 28.43 & 13.27 & 71.67 & 27.00 & 62.33 \\
 & 6000  & 22.50 &  7.50 & 66.67 & 31.10 & 29.43 &  5.37 & 74.33 & 28.00 & 62.33 \\
 & 10000 & \textbf{28.34} & \textbf{7.50} & \textbf{73.54} & 33.11 & 29.10 & 12.11 & \textbf{76.33} & \textbf{26.67} & \textbf{65.06} \\
 & 13000 & 17.50 & 12.50 & 28.57 & \textbf{32.44} & \textbf{26.67} & \textbf{17.79} & 73.33 & 26.00 & 64.54 \\
 & 16000 & 17.50 &  7.50 & 57.14 & 32.11 & 30.10 &  6.26 & 75.00 & 29.33 & 60.89 \\
\bottomrule
\end{tabular}
\end{table*}

\noindent \textbf{Impact of Search Scope} We vary the size of the initial candidate pool (search scope) from 2,000 to 16,000. Since random sampling under different beam widths causes the value of $\mathcal{A}_{clean}$ to vary and subsequently affect $\mathcal{A}_{attack}$, ASR normalizes this fluctuation and serves as our key evaluation metric. Table~\ref{tab:beam_width_trend_2} indicates that a scope of 13,000 generally yields the best trade-off between ASR and efficiency. Smaller scopes (e.g., 2,000) limit candidate diversity, reducing effectiveness; larger scopes (16,000) increase computation with diminishing returns. This demonstrates AGLS's ability to effectively utilize a bounded search space.

\subsection{Discussion}
\label{sec:discussion}
The experimental results collectively demonstrate that AGLS is a highly effective and practical attack against LLMs equipped with visible prompt optimizers. Its superiority stems from three design principles:
\begin{itemize}
    \item \textbf{Constraint-Aware Search:} By explicitly modeling and enforcing semantic similarity constraints at each generation step, AGLS avoids the “semantic collapse” common in one-shot attacks, producing more natural adversarial examples.
    \item \textbf{Adaptive Feedback:} The dynamic adjustment mechanism allows AGLS to recover from suboptimal local decisions, mitigating the early-commitment bias inherent in static beam search methods.
    \item \textbf{Efficient Black-Box Operation:} AGLS achieves state-of-the-art performance without requiring model gradients, massive query budgets, or white-box access, making it directly applicable to real-world commercial systems.
\end{itemize}
The lower ASR on textual QA tasks suggests potential avenues for improvement, such as incorporating discourse-level constraints or leveraging stronger sentence encoders. Nevertheless, AGLS establishes a strong baseline for semantic-preserving hijacking attacks and underscores the urgent need to harden prompt optimization modules against such adversarial manipulation.

\section{Conclusion}

This paper introduces Adaptive Greedy Local Search, an effective adversarial attack targeting visible prompt optimizers in LLMs. By simulating the multi-stage rewriting process and incorporating adaptive semantic feedback, AGLS crafts perturbations that preserve meaning while misleading model behavior. Extensive experiments demonstrate AGLS's superior attack success rates compared to existing semantic-preserving methods. Our work reveals a critical vulnerability in transparency-enhancing optimization modules and underscores the need for security-aware design in deployable LLM systems. Future work will extend AGLS to multimodal and multi-turn settings and explore corresponding defense mechanisms.

The AGLS method also has limitations in some cases, which can be targeted to defend against similar attacks. (1) This kind of attack can be identified by multi-index consistency detection; (2) Introducing a randomization mechanism in the candidate content screening step of the optimizer to effectively resist the attack; (3) If the core keywords are wrongly identified in the attack, the attack effect of perturbation will be greatly reduced.

\section{Acknowledgment}

This work was partially supported by the Research Development Fund with No. Grant No. RDF-22-01-020 and the “Qing Lan Project” in Jiangsu universities.           

\clearpage

\bibliographystyle{IEEEbib}
\bibliography{icme2026references}

\clearpage

\appendix

\section{Dataset Details and Evaluation Protocol}
\label{appsec:detail-dataset}

\subsection{Datasets Details}

\noindent \textbf{GSM8K \cite{cobbe2021training}:} A collection of linguistically diverse grade-school math word problems that require multi-step reasoning. The dataset contains 8,000 problems with corresponding step-by-step solutions, making it suitable for evaluating mathematical reasoning under adversarial conditions.

\noindent \textbf{Math QA \cite{amini2019mathqa}:} A dataset of math word problems annotated with operation-based formalisms, explanations, and multiple-choice options. It emphasizes interpretable problem-solving and is used to assess numerical reasoning robustness.

\noindent \textbf{Strategy QA \cite{geva2021strategyqa}:} A question-answering benchmark requiring implicit multi-hop reasoning, where answers are not directly extractable from the provided context, thereby evaluating deeper comprehension and inference skills.

\noindent \textbf{SVAMP \cite{patel2021svamp}:} A question-answering dataset specifically designed to test the ability of models to solve simple math word problems, often used to evaluate generalization and robustness.

\noindent \textbf{SQuAD \cite{rajpurkar2016squad}:} The Stanford Question Answering Dataset contains over 100,000 question-answer pairs derived from Wikipedia articles. It is widely used for evaluating reading comprehension and extractive QA capabilities.

\noindent \textbf{SQuAD 2.0 \cite{rajpurkar2018know}:} An extended version of SQuAD that introduces unanswerable questions, challenging models to distinguish between answerable and unanswerable queries, thus testing reasoning and discrimination abilities.

\subsection{Victim Models Details}
\label{appsec:detail-victim}

\noindent Our experiments encompass both commercial and open-source large language models to ensure broad evaluation coverage:
\\

\noindent \textbf{GPT-4 Series \cite{radford2020chatgpt}:} We test GPT-4-Turbo and GPT-4o, two of the most capable commercial models developed by OpenAI, known for their strong reasoning and instruction-following abilities.

\noindent \textbf{Llama Series \cite{dubey2024llama}} We include multiple versions of Meta’s open-source models: Llama-3.1-8B, Llama-3.1-70B, Llama-3.2-3B, and Llama-3.3-70B, covering a range of parameter scales and architectural variations.

\noindent \textbf{Qwen2.5 Series \cite{yang2024qwen2}:} We evaluate Alibaba’s open-source models Qwen2.5-0.5B, Qwen2.5-7B, and Qwen2.5-14B, which are known for strong multilingual and reasoning performance.

\noindent \textbf{Gemma2 \cite{team2024gemma}:} We test Google’s efficient open models Gemma2-2B and Gemma2-9B, designed for high performance at practical model sizes.

\subsection{Evaluation Criteria}
\label{appsec:detail-add}
We adopt distinct correctness criteria for textual and numerical responses to accurately reflect task requirements:

\noindent \textbf{Textual Responses:} Let $A$ be the set of words in the model’s response and  $B$ the set of words in the ground-truth answer

\begin{itemize}
    \item if $|B| \ge 3$: The response is considered correct if $|A \cap B| > 2$.
    \item if $|B| < 3$: The response is correct only if $B \subset A$.
\end{itemize}

\noindent \textbf{Numerical Responses:} A response is considered correct if the extracted numerical value exactly matches the ground-truth answer. Any deviation, including formatting differences, is treated as incorrect.

\subsection{Prompt Templates}
\label{appendix: prompts}

\subsubsection{Ollama text responses prompts}

\begin{Verbatim}[frame=single, breaklines=true, breaksymbol={}]
Please give me a brief answer directly and promise to answer in English:
\end{Verbatim}

\subsubsection{Ollama numerical responses prompts}

\begin{Verbatim}[frame=single, breaklines=true, breaksymbol={}]
Give me the numerical answers directly, without giving the intermediate steps:
\end{Verbatim}

\subsubsection{OpenAI text responses prompts}

\begin{Verbatim}[frame=single, breaklines=true, breaksymbol={}]
Please give me a brief answer directly to the following questions and promise to answer in English:
\end{Verbatim}

\subsubsection{OpenAI numerical responses prompts}

\begin{Verbatim}[frame=single, breaklines=true, breaksymbol={}]
Give me the numerical answers directly in the following questions, without giving the intermediate steps:
\end{Verbatim}

\section{Analysis of Semantic Similarity Dynamics in AGLS}

\subsection{Semantic Trajectories During Attack Generation}

\begin{figure}
    \centering
    \includegraphics[width=1.0\linewidth]{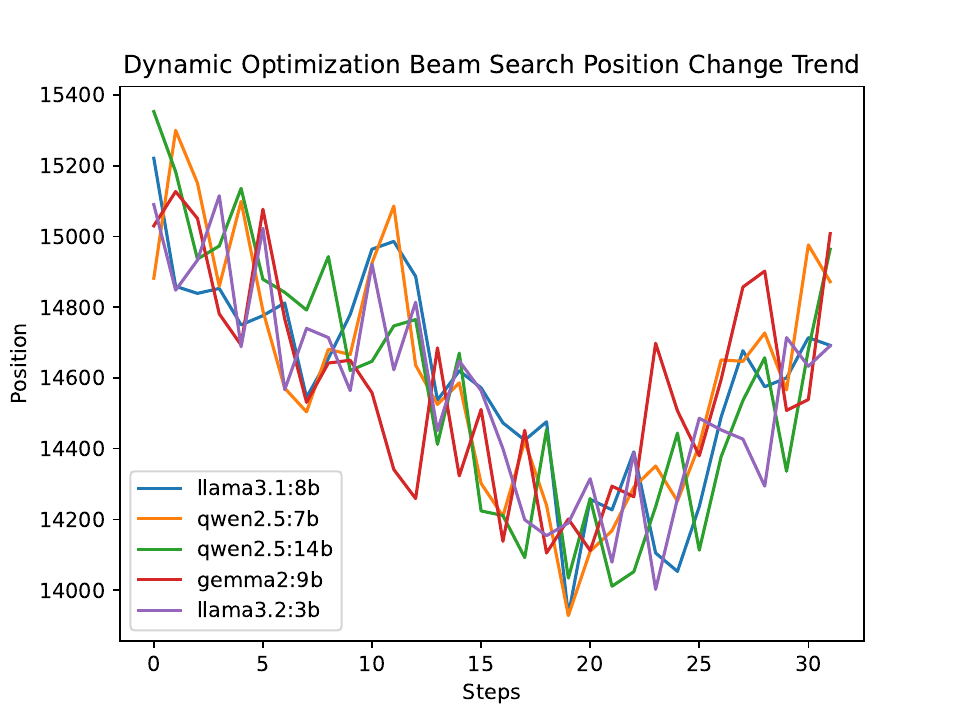}
    \caption{AGLS position change trend}
    \label{fig:position-trend}
\end{figure}

\begin{figure}
    \centering
    \includegraphics[width=1.0\linewidth]{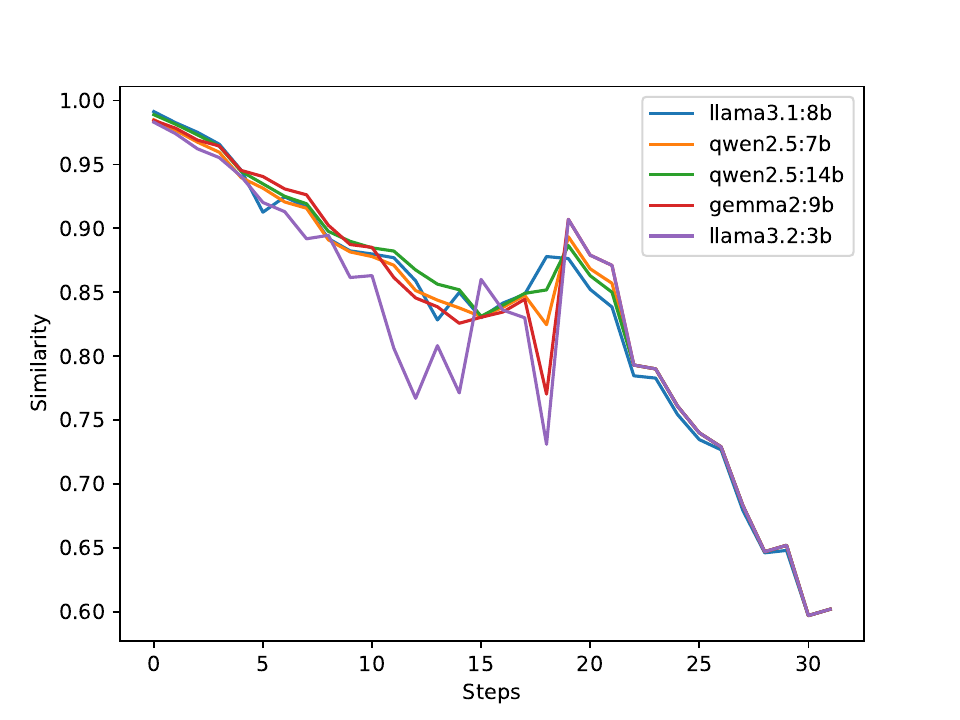}
    \caption{AGLS semantic similarity change trend}
    \label{fig:similarity-trend}
\end{figure}

To understand how AGLS maintains semantic coherence while introducing adversarial perturbations, we analyze the similarity trajectories during the generation of attack samples. Figure \ref{fig:position-trend} and Figure \ref{fig:similarity-trend} illustrate the positional adjustment trends and semantic similarity changes, respectively, when attacking the same dataset (GSM8K) across different victim models. Key Observations are as follows:

\textbf{Consistent Adjustment Patterns:} Despite differences in model architecture and scale, AGLS exhibits similar positional adjustment trends (Figure \ref{fig:position-trend}). Inflection points in candidate selection occur at roughly the same generation steps, indicating that the adaptive feedback mechanism operates consistently across models.

\textbf{Controlled Semantic Drift:} Figure \ref{fig:similarity-trend} shows that semantic similarity follows a predictable trajectory, with inflection points typically occurring around steps 18–19. While the magnitude of similarity variation differs across models (e.g., larger changes in Llama3.2-3B and Gemma2-9B), the overall trend remains stable, confirming that AGLS effectively balances semantic preservation and adversarial intent.

\subsection{Search Scope Analysis}

The search scope—defined as the size of the candidate pool considered at each step—directly influences AGLS’s efficiency and effectiveness. Table~\ref{tab:beam_width_trend_1} and Figures 3–6 present a comprehensive analysis of how varying the search scope affects the Attack Success Rate (ASR) across models and datasets. The key findings are listed as follows:

\textbf{Optimal Scope:} A search scope of 13,000 candidates generally yields the highest ASR while maintaining reasonable computational cost. This scope provides sufficient lexical diversity without introducing excessive noise.

\textbf{Scope-Performance Relationship:} Narrower scopes (e.g., 2,000) limit candidate diversity and reduce ASR, while excessively broad scopes (e.g., 16,000) offer diminishing returns and increase inference time.

\textbf{Model-Specific Behavior:} Some models, such as Qwen2.5-14B, achieve peak performance at a scope of 10,000 on certain datasets (GSM8K, SVAMP), suggesting that the optimal scope may depend on model architecture and task type.

\subsection{Additional Attack Results}

Table \ref{tab:combined-attack} presents extended experimental results for models not included in the main paper due to space constraints. Key takeaways include:

\textbf{Smaller Models Remain Highly Vulnerable:} Even very small models such as Qwen2.5-0.5B and Gemma2-2B exhibit high ASR on numerical tasks (e.g., 92.31\% on GSM8K for Qwen2.5-0.5B), indicating that model scale alone does not confer robustness against semantic hijacking.

\textbf{Consistent Trends Across Tasks:} The relative performance of AGLS remains stable across textual and numerical tasks, reinforcing its generality as an attack framework.

\subsection{Qualitative Examples}

Below we provide a concrete example illustrating how AGLS subtly alters an input question while preserving its surface semantics.

\noindent \textbf{Original Question (GSM8K):}

\begin{tcolorbox}[colback=white, colframe=black, boxrule=1pt, arc=0mm, breakable]
Isabelle works in a hotel and runs a bubble bath for each customer who enters the hotel. There are 13 rooms for couples and 14 single rooms. For each bath that is run, Isabelle needs 10ml of bubble bath. If every room is filled to maximum capacity, how much bubble bath, in millilitres, does Isabelle need?
\end{tcolorbox}

\noindent \textbf{AGLS-Perturbed Question:}

\begin{tcolorbox}[colback=white, colframe=black, boxrule=1pt, arc=0mm, breakable]
Isabelle \textcolor{blue}{is employed} in a hotel and \textcolor{blue}{prepares} a bubble bath for each customer who \textcolor{blue}{arrives} at the hotel. There \textcolor{blue}{exist} 13 rooms for couples and 14 single \textcolor{blue}{accommodations}. For each bath that is prepared, Isabelle \textcolor{blue}{requires} 10ml of bubble bath. If every room is \textcolor{blue}{occupied} to maximum capacity, how much bubble bath, in millilitres, \textcolor{blue}{does} Isabelle \textcolor{blue}{require}?
\end{tcolorbox}

The perturbed version substitutes verbs (``works" $>$ ``is employed", ''runs" $>$ ``prepares") and nouns (``rooms" $>$ ``accommodations") with semantically similar alternatives. These changes are grammatically sound and preserve the original problem structure, yet they are sufficient to mislead certain LLMs into producing incorrect answers, demonstrating the stealth and effectiveness of AGLS.

\section{Implementation and Reproducibility}

All experiments were conducted in a strictly black-box setting. We used \textbf{bert-base-uncased} as the backbone model for candidate generation and similarity computation. Hyperparameters were fixed as follows: semantic similarity threshold $\sigma=0.80$ (BERTScore F1), candidate pool size = 13,000, beam width = 4, and adaptive step size $s = 50$. Code, detailed configuration files, and a subset of generated adversarial examples are available at our anonymized repository: \url{https://anonymous.4open.science/r/EDA5G6S8}.


\begin{table*}[ht]
\centering
\footnotesize
\caption{The result of attack success rate via the AGLS search scope changes experiments (Llama3.2-3B/8B, Qwen2.5-7B/14B).}
\label{tab:beam_width_trend_1}
\begin{tabular}{l|c|ccccccccccc}
\toprule
\multirow{2}{*}{\textbf{Models}} & \multirow{2}{*}{\textbf{Search scope}} & \multicolumn{3}{c}{\textbf{GSM8K}} & \multicolumn{3}{c}{\textbf{SQUAD}} & \multicolumn{3}{c}{\textbf{SVAMP}}\\
\cmidrule(lr){3-5} \cmidrule(lr){6-8} \cmidrule(lr){9-11}
 &  & $\mathcal{A}_{\textrm{clean}}$ & $\mathcal{A}_{\textrm{attack}}$ & $\textbf{ASR}$ $\uparrow$ & $\mathcal{A}_{\textrm{clean}}$ & $\mathcal{A}_{\textrm{attack}}$ & $\textbf{ASR}$ $\uparrow$ & $\mathcal{A}_{\textrm{clean}}$ & $\mathcal{A}_{\textrm{attack}}$ & $\textbf{ASR}$ $\uparrow$ \\
\midrule
 & 2000 & 55.00 & 10.00 & 81.82 & 26.76 & 22.07 & 17.53 & 43.00 & 8.67 & 79.84\\
 & 6000 & 47.50 & 2.50 & 94.74 & 26.09 & 23.08 & 11.54 & 40.33 & 7.00 & 82.64 \\
\textbf{Llama3.2-3B}   & 10000 & 25.00 & 2.50 & 90.00 & 25.08 & 21.74 & 13.32 & 38.00 & 7.33 & 80.71 \\
& 13000 & \textbf{42.50} & \textbf{2.50} & \textbf{94.12} & \textbf{25.42} & \textbf{17.48} & \textbf{31.24} & \textbf{39.00} & \textbf{6.00} & \textbf{84.62} \\
& 16000 & 40.00 & 2.50 & 93.75 & 25.08 & 21.40 & 14.67 & 36.00 & 7.33 & 79.64 \\
\hline
  & 2000 & 27.50 & 12.50 & 54.54 & 32.11 & 29.10 & 9.37 & 20.33 & 10.67 & 47.52 \\
  & 6000 & 20.00 & 5.00 & 75.00 & 31.10 & 26.76 & 13.95 & \textbf{18.33} & \textbf{6.33} & \textbf{65.47} \\
\textbf{Llama3.1-8B}   & 10000 & 22.50 & 2.50 & 88.89 & 31.10 & 29.43 & 11.11 & 18.33 & 10.67 & 41.79 \\
  & 13000 & \textbf{25.00} & \textbf{2.50} & \textbf{90.00} & \textbf{35.45}  & \textbf{29.77} & \textbf{16.02} & 16.33 & 10.00 & 38.76 \\
  & 16000 & 17.50 & 7.50 & 57.14 & 34.11 & 28.76 & 15.68 & 17.00 & 10.67 & 37.24 \\
\hline
  & 2000 & 15.00 & 10.00 & 33.33 & 20.40 & 17.73 & 13.09 & 53.33 & 20.00 & 62.50 \\
  & 6000 & 10.00 & 6.25 & 37.50 & 20.74 & 18.73 & 9.69 & 54.33 & 18.67 & 65.64 \\
  \textbf{Qwen2.5-7B}    & 10000 & 17.50 & 10.00 & 42.86 & 20.74 & 19.73 & 4.87 & 54.67 & 19.67 & 64.02 \\
  & 13000 & \textbf{15.00} & \textbf{7.50} & \textbf{50.00} & \textbf{20.40} & \textbf{14.36} & \textbf{29.61} & \textbf{54.00} & \textbf{16.45} & \textbf{69.54} \\
  & 16000 & 10.00 & 7.50 & 25.00 & 19.40 & 18.39 & 5.21 & 52.67 & 22.33 & 62.96 \\
  \hline
    & 2000 & 10.00 & 7.50 & 2.50 & 32.78 & 28.43 & 13.27 & 71.67 & 27.00 & 62.33 \\
    & 6000 & 22.50 & 7.50 & 66.67 & 31.10 & 29.43 & 5.37 & 74.33 & 28.00 & 62.33 \\
    \textbf{Qwen2.5-14B} & 10000 & \textbf{28.34} & \textbf{7.50} & \textbf{73.54} & 33.11 & 29.10 & 12.11 & \textbf{76.33} & \textbf{26.67} & \textbf{65.06} \\
    & 13000 & 17.50 & 12.50 & 28.57 & \textbf{32.44} & \textbf{26.67} & \textbf{17.79} & 73.33 & 26.00 & 64.54 \\
    & 16000 & 17.50 & 7.50 & 57.14 & 32.11 & 30.10 & 6.26 & 75.00 & 29.33 & 
    60.89 \\
\bottomrule
\end{tabular}
\end{table*}

\begin{figure}[ht]
    \centering
    \begin{subfigure}[b]{0.48\textwidth}
        \includegraphics[width=\textwidth]{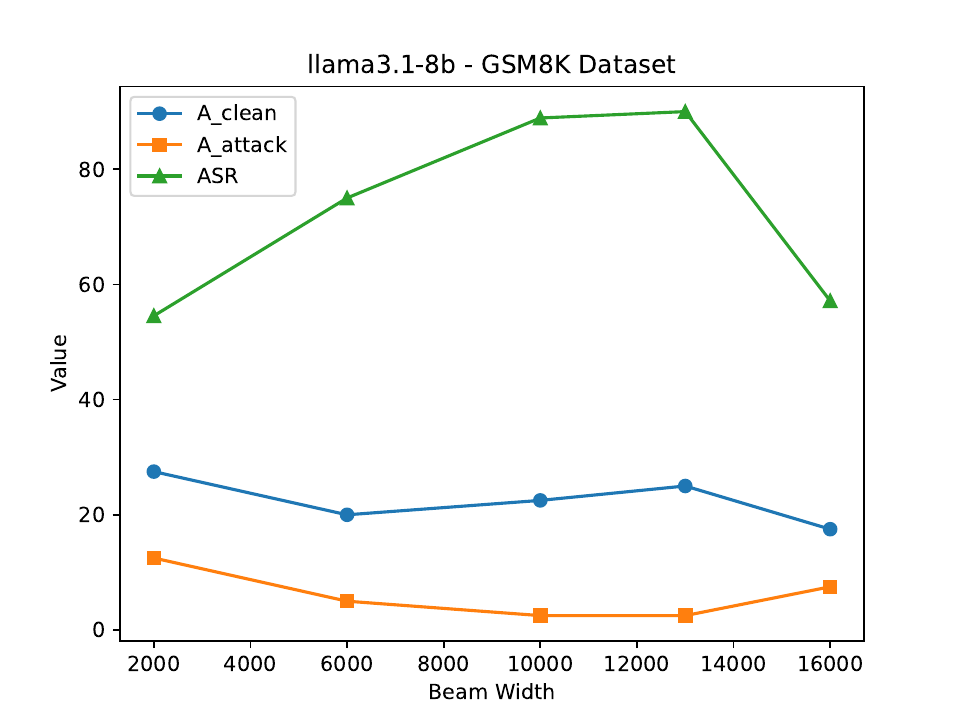}
        \caption{GSM8K on llama3.1-8B}
    \end{subfigure}
    \begin{subfigure}[b]{0.48\textwidth}
        \includegraphics[width=\textwidth]{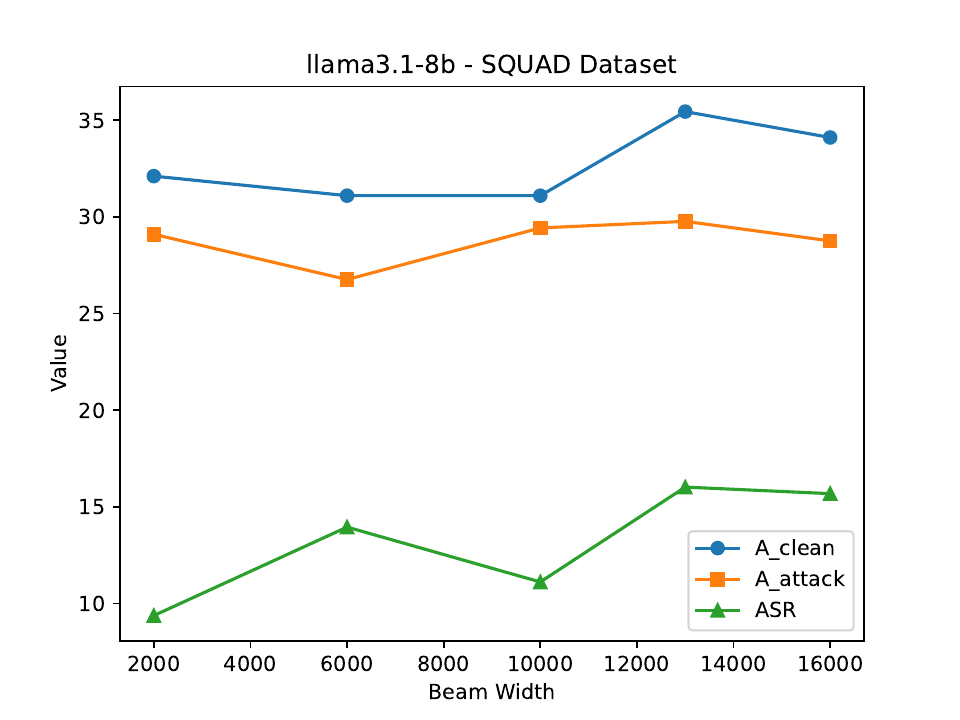}
        \caption{SQUAD on llama3.1-8B}
    \end{subfigure}
    \begin{subfigure}[b]{0.48\textwidth}
        \includegraphics[width=\textwidth]{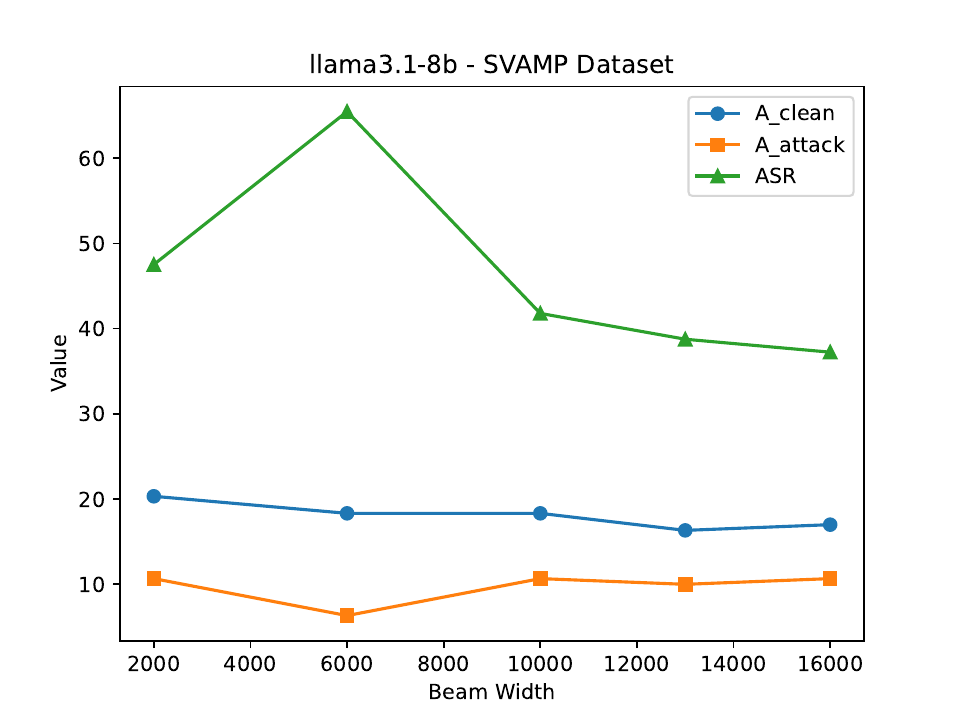}
        \caption{SVAMP on llama3.1-8B}
    \end{subfigure}
    \caption{The relationship between the AGLS search scope and the attack success rate. (Part I)}
    \label{fig:beamwidth_asr_1}
\end{figure}

\begin{figure}[ht]
    \centering
    \begin{subfigure}[b]{0.48\textwidth}
        \includegraphics[width=\textwidth]{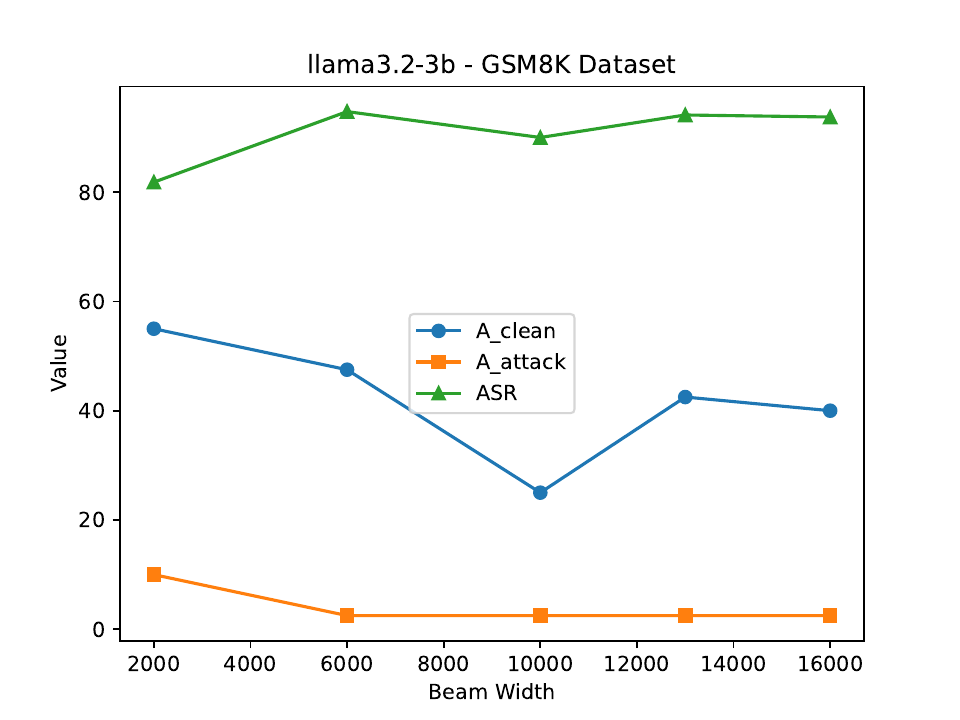}
        \caption{GSM8K on llama3.2-3B}
    \end{subfigure}
    \begin{subfigure}[b]{0.48\textwidth}
        \includegraphics[width=\textwidth]{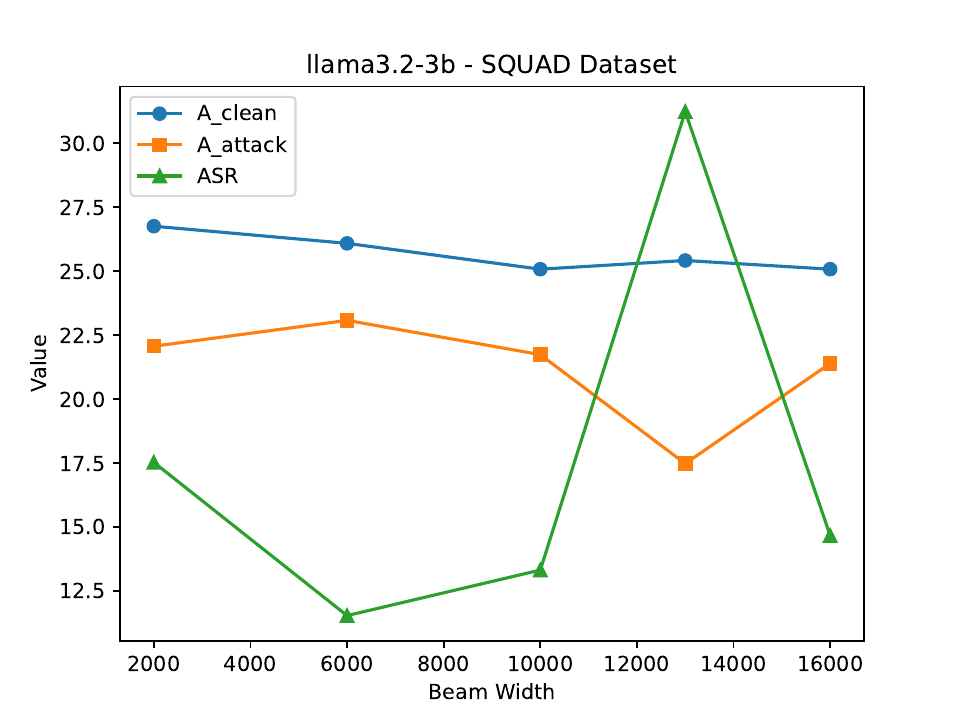}
        \caption{SQUAD on llama3.2-3B}
    \end{subfigure}
    \begin{subfigure}[b]{0.47\textwidth}
        \includegraphics[width=\textwidth]{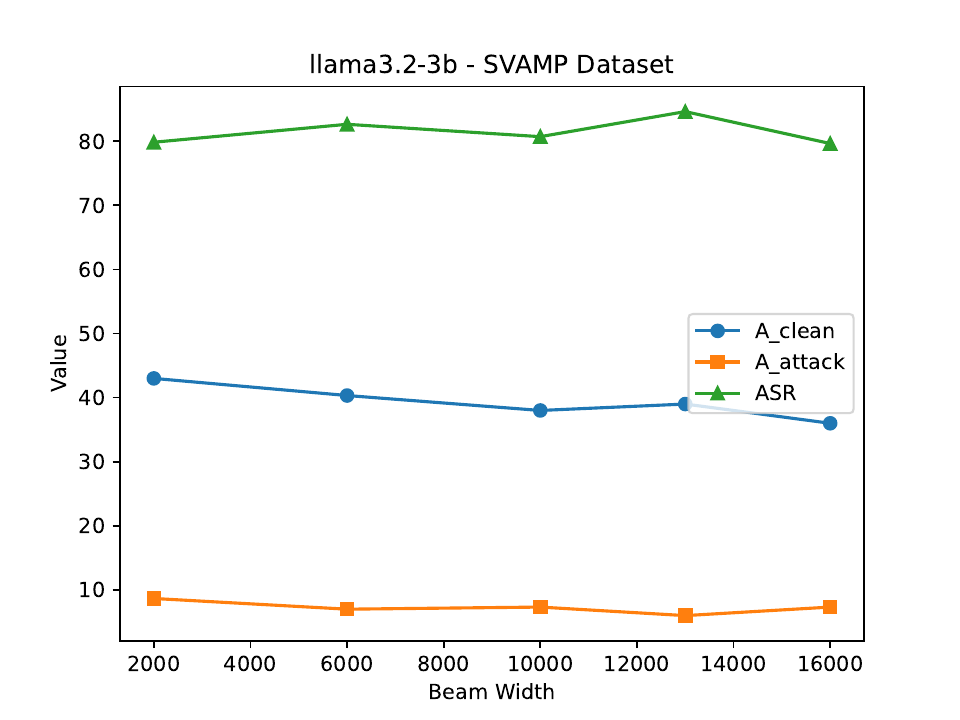}
        \caption{SVAMP on llama3.2-3B}
    \end{subfigure}
    \caption{The relationship between the AGLS search scope and the attack success rate. (Part II)}
    \label{fig:beamwidth_asr_2}
\end{figure}

\begin{figure}[!ht]
    \centering
    \begin{subfigure}[b]{0.48\textwidth}
    \includegraphics[width=\textwidth]{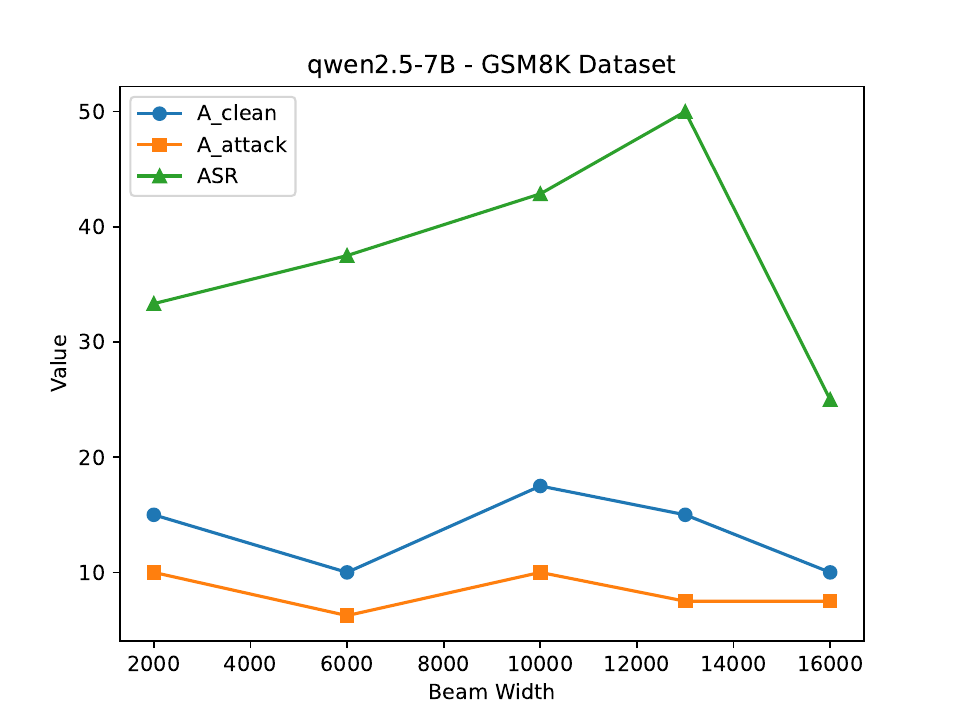}
        \caption{GSM8K on qwen2.5-7B}
    \end{subfigure}
    \hfill 
    \begin{subfigure}[b]{0.48\textwidth}
        \includegraphics[width=\textwidth]{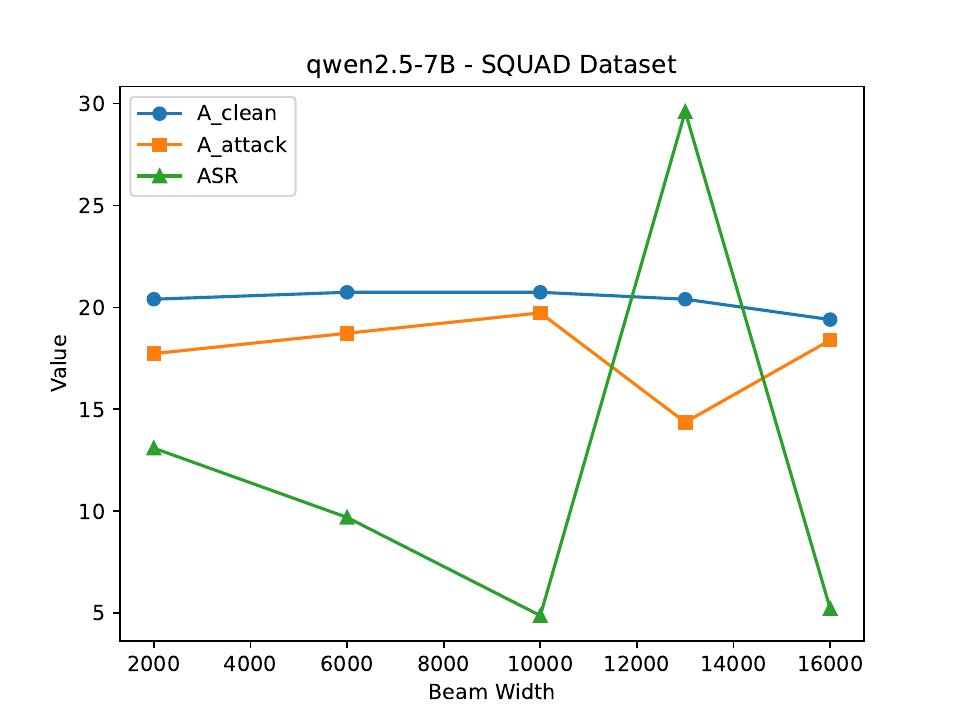}
        \caption{SQUAD on qwen2.5-7B}
    \end{subfigure}
    \vspace{10pt} 
    \begin{subfigure}[b]{0.48\textwidth}
        \includegraphics[width=\textwidth]{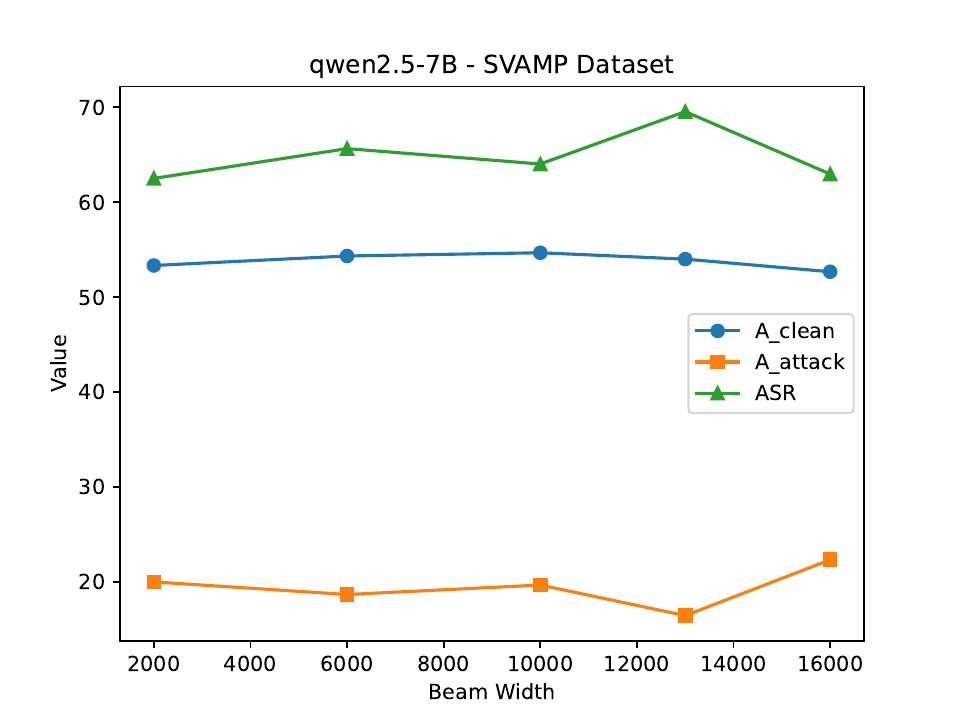}
        \caption{SVAMP on qwen2.5-7B}
    \end{subfigure}
    \caption{The relationship between the AGLS search scope and the attack success rate. (Part III)}
    \label{fig:beamwidth_asr_3}
\end{figure}

\begin{figure}[!ht]
    \begin{subfigure}[b]{0.48\textwidth}
        \includegraphics[width=\textwidth]{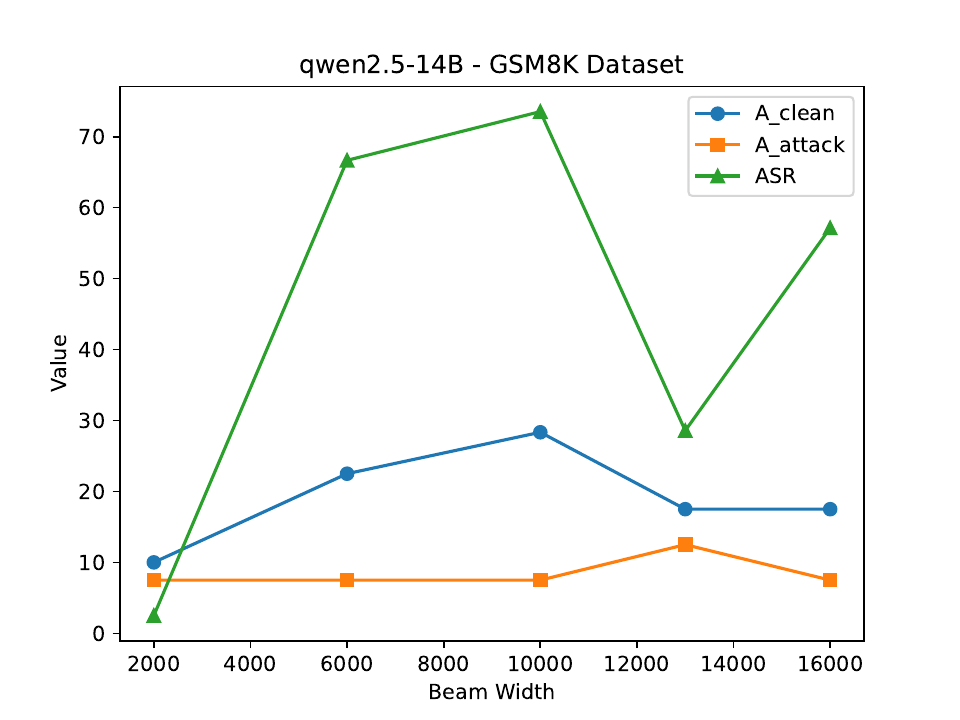}
        \caption{GSM8K on qwen2.5-14B}
    \end{subfigure}
    \vspace{10pt}
    \begin{subfigure}[b]{0.48\textwidth}
        \includegraphics[width=\textwidth]{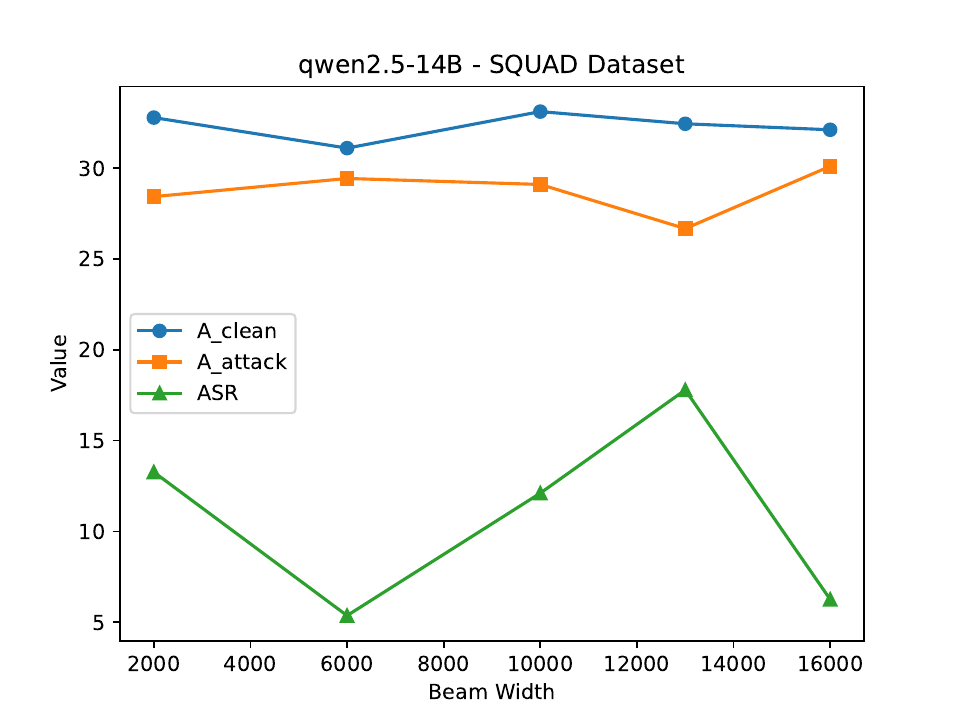}
        \caption{SQUAD on qwen2.5-14B}
    \end{subfigure}
    \hfill
    \begin{subfigure}[b]{0.48\textwidth}
        \includegraphics[width=\textwidth]{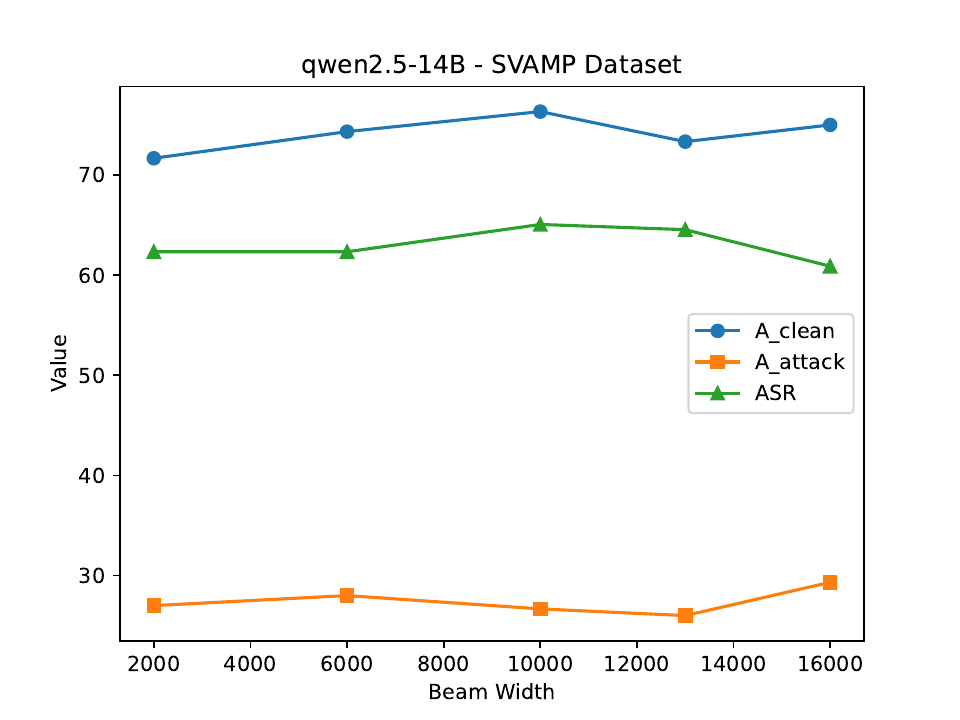}
        \caption{SVAMP on qwen2.5-14B}
    \end{subfigure}
    \caption{The relationship between the AGLS search scope and the attack success rate. (Part IV)}
    \label{fig:beamwidth_asr_4}
\end{figure}

\begin{table*}[!ht]
\centering
\footnotesize
\caption{Comparison of attack effects of AGLS on different LLMs and datasets}
\label{tab:combined-attack}
\begin{tabular}{lcccccccc}
\toprule
\multirow{2}{*}{\textbf{Models}} & \multicolumn{4}{c}{\textbf{GSM8K}} & \multicolumn{4}{c}{\textbf{Math QA}} \\
\cmidrule(lr){2-5} \cmidrule(lr){6-9}
 & $A_{\textrm{clean}}$ & $A_{\textrm{attack}}$ & \textbf{ASR}↑ & $\text{T}_{\text{avg}}$ 
 & $A_{\textrm{clean}}$ & $A_{\textrm{attack}}$ & \textbf{ASR}↑ & $\text{T}_{\text{avg}}$ \\ 
\midrule
llama3.1-70B & 47.50 & 15.00 & \cellcolor[gray]{0.9}68.42 & 3.44 & 13.67 & 11.33 & \cellcolor[gray]{0.9}17.12 & 0.14 \\
llama3.3-70B & 72.50 & 17.50 & \cellcolor[gray]{0.9}75.86 & 9.45 & 32.33 & 23.67 & \cellcolor[gray]{0.9}26.79 & 1.45 \\
\hline
qwen2.5-0.5B & 32.50 & 2.50 & \cellcolor[gray]{0.9} 92.31 & 1.84 & 6.00 & 2.33 & \cellcolor[gray]{0.9}61.17 & 1.81 \\
gemma2-2B & 50.00 & 5.00 & \cellcolor[gray]{0.9} 90.00 & 1.49 & 46.72 & 34.79 & \cellcolor[gray]{0.9}25.54 & 0.38 \\
\bottomrule
\end{tabular}
\vspace{0.5em}
\begin{tabular}{lcccccccc}
\toprule
\multirow{2}{*}{\textbf{Models}} & \multicolumn{4}{c}{\textbf{SQUAD}} & \multicolumn{4}{c}{\textbf{Strategy QA}} \\
\cmidrule(lr){2-5} \cmidrule(lr){6-9}
 & $A_{\textrm{clean}}$ & $A_{\textrm{attack}}$ & \textbf{ASR}↑ & $\text{T}_{\text{avg}}$ 
 & $A_{\textrm{clean}}$ & $A_{\textrm{attack}}$ & \textbf{ASR}↑ & $\text{T}_{\text{avg}}$ \\ 
\midrule
llama3.1-70B & 49.83 & 44.15 & \cellcolor[gray]{0.9}11.40 & 2.01s & 49.67 & 47.00 & \cellcolor[gray]{0.9}5.38 & 2.03s \\
llama3.3-70B & 52.17 & 49.16 & \cellcolor[gray]{0.9}5.77 & 2.53s & 50.33 & 47.33 & \cellcolor[gray]{0.9}5.96 & 1.49 \\
\hline
qwen2.5-0.5B & 9.03 & 8.03 & \cellcolor[gray]{0.9}11.07 & 0.64 & 26.67 & 25.33 & \cellcolor[gray]{0.9}5.02 & 1.09s \\
gemma2-2B & 19.40 & 15.97 & \cellcolor[gray]{0.9}17.68 & 0.38 & 51.67 & 36.67 & \cellcolor[gray]{0.9}29.03 & 0.26 \\
\bottomrule
\end{tabular}
\end{table*}

\end{document}